\documentclass[final]{cvpr}
\usepackage{appendix}
\usepackage{times}
\usepackage{epsfig}
\usepackage{graphicx}
\usepackage{amsmath}
\usepackage{amssymb}
\usepackage{booktabs}
\usepackage{multirow}
\usepackage{caption}
\usepackage{array}
\usepackage[pagebackref=true,breaklinks=true,colorlinks,bookmarks=false]{hyperref}
\usepackage{multirow}

\def\cvprPaperID{7907} % *** Enter the CVPR Paper ID here
\def\confYear{CVPR 2022}

\begin{document}
\captionsetup[figure]{labelfont={bf},labelformat={default},labelsep=period,name={Fig.}}
\title{Underwater Light Field Retention : Neural Rendering for Underwater Imaging}
% \author{Tian Ye$^{\dag}$$^{1}$, Sixiang Chen$^{\dag}$$^{1}$, Yun Liu$^{2}$,
% Yi Ye$^{1}$,
% Erkang Chen$^{1}$\thanks{Corresponding author.$^{\dag}$Equal contribution.}$_$,
% Yuche Li$^{3}$\\
% $^1$ School of Ocean Information Engineering,Jimei University, Xiamen, China\\
% $^2$ College of Artificial Intelligence, Southwest University, Chonqing, China\\
% $^3$ College of Geosciences, China University of Petroleum, Beijing, China\\
% \small  $\left\{ 201921114031, 201921114013 \right\}$ \small @jmu.edu.cn \\
% \small yunliu@swu.edu.cn, 201921114003@jmu.edu.cn, ekchen@jmu.edu.cn, liyuche$\_$cq@163.com }

\author{Tian Ye$^{\dag 1}$, Sixiang Chen$^{\dag 1}$, Yun Liu$^{2}$, Yi Ye$^{1}$, Erkang Chen$^{1}$\thanks{Corresponding author. $^\dag$Equal contribution.}, , Yuche Li$^{3}$\\
$^1$ School of Ocean Information Engineering,Jimei University, Xiamen, China\\
$^2$ College of Artificial Intelligence, Southwest University, Chonqing, China\\
$^3$ College of Geosciences, China University of Petroleum, Beijing, China\\
\small  $\left\{ 201921114031, 201921114013 \right\}$ \small @jmu.edu.cn \\
\small yunliu@swu.edu.cn, 201921114003@jmu.edu.cn, ekchen@jmu.edu.cn, liyuche$\_$cq@163.com }

% \institute{$^1$School of Ocean Information Engineering,Jimei University, Xiamen, China\\
%     $^2$Southwest University School of Artificial Intelligence, Chonqing, China}
% \and
% \\
% \\
% \author{ Sixiang\and\\
%  \\
% Institution1 address\\}
\maketitle

%%%%%%%%% ABSTRACT
\begin{abstract}
Underwater Image Rendering aims to generate a true-to-life underwater image from a given clean one, which could be applied to various practical applications such as underwater image enhancement, camera filter, and virtual gaming. We explore two less-touched but challenging problems in underwater image rendering, namely, i) how to render diverse underwater scenes by a single neural network? ii) how to adaptively learn the underwater light fields from natural exemplars, \textit{i,e.}, realistic underwater images? To this end, we propose a neural rendering method for underwater imaging, dubbed UWNR (Underwater Neural Rendering). Specifically, UWNR is a data-driven neural network that implicitly learns the natural degenerated model from authentic underwater images, avoiding introducing erroneous biases by hand-craft imaging models. 

Compared with existing underwater image generation methods, UWNR utilizes the natural light field to simulate the main characteristics of the underwater scene. Thus, it is able to synthesize a wide variety of underwater images from one clean image with various realistic underwater images. 

Extensive experiments demonstrate that our approach achieves better visual effects and quantitative metrics over previous methods. Moreover, we adopt UWNR to build an open Large Neural Rendering Underwater Dataset containing various types of water quality, dubbed LNRUD. The source code and LNRUD are
available at \url{https://github.com/Ephemeral182/UWNR}.  
\let\thefootnote\relax\footnote{This work was supported in part by the Natural Science Foundation of Fujian Province of China (2021J01867), Education Department of Fujian Province (JAT190301), Foundation of Jimei University (ZP2020034), Natural Science Foundation of Chongqing, China (cstc2020jcyj-msxmX0324), the Construction of Chengdu-Chongqing Economic Circle Science and Technology Innovation Project (KJCX2020007).}
 \end{abstract}

%%%%%%%%% BODY TEXT
\begin{figure}[!t]
\vspace{-0.3cm}
\setlength{\abovecaptionskip}{0.1cm} %调整caption与图的距离
\setlength{\belowcaptionskip}{-0.4cm}%调整caption与下文的距离
\centering 
\includegraphics[width=7cm]{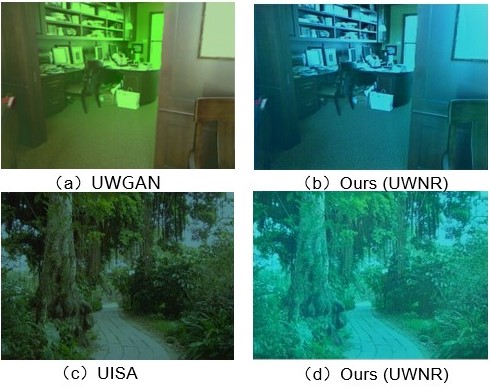}

\caption{Visual comparison of our method and previous underwater rendering methods. (a) and (c) are respectively obtained by UWGAN \cite{wang2021uwgan} and UISA \cite{hou2020benchmarking}. (b) and (d) are rendered by our method. The underwater rendering results by our method have more natural visual effects closing to real underwater scenes.}
\label{fig:0}
\end{figure}
\section{Introduction}
\label{sec:intro}
The underwater environment and resource has been
gradually developed and explored in the last decade~\cite{bailey2008archaeology,gu2018overview,komatsu2018optical,eldred2021design,jahanbakht2021internet}. Unlike the terrestrial environment, it is hard for us to obtain satisfactory paired underwater datasets, limiting the development of many underwater vision tasks. For example, underwater image enhancement is a practical computer vision task that requires paired underwater images with ideal ground-truth for training. However, manual collection often requires a lot of workforce and material resources.~\cite{li2019underwater,islam2020simultaneous,islam2020fast}. Furthermore, the visual effect of the ground-truth generated by various algorithms in current underwater image datasets is not satisfactory for implementation. And the synthetic underwater datasets based on the traditional underwater imaging model hardly model the complex degeneration of the underwater environment well. In particular, underwater image rendering that effectively generates true-to-life underwater images becomes a novel and valuable technique compared to the high cost of acquiring paired underwater datasets with real underwater images.

Previous underwater image restoration works~\cite{drews2013transmission,anwar2018deep} usually utilize the following formula as the underwater imaging model:
\begin{equation}\label{eq1}
    \mathcal{I}(x) = \mathcal{J}(x)t(x)+{B}(x)(1-t(x)).
\end{equation}
where $B(x)$ is the background light and $\mathcal{J}(x)$ is the image without degrading by underwater particle scattering. $t(x)=e^{-\beta d(x)}$ is the transmission map, $\beta$ is the scattering coefficient and $d(x)$ is the depth of scene. The $\mathcal{I}(x)$ is the underwater image.

Image generation like image rendering etc. usually involves adversarial generative networks (GAN) \cite{goodfellow2014generative,Zhu_2017_ICCV,karras2019style,karras2020analyzing,li2021weather}. Zhu $et$ $al.$  \cite{Zhu_2017_ICCV} utilized a cycle-consistent GAN to implement image style transfer. Proposing mapping networks and adaptive Instance Normalization techniques were proposed in a large generative network Stylegan\cite{karras2019style,karras2020analyzing}. Li $et$ $al.$ \cite{li2021weather} presented Weather GAN to render different weather scenes and migrated weather conditions from one category to another. The above GAN-based approaches generate impressive results.
However, there are well-known shortcomings for GAN methods. For example, GAN-based methods are prone to mode collapse and create fake features.

\begin{figure*}[!t]
\vspace{-0.3cm}
\setlength{\abovecaptionskip}{0.0cm} %调整caption与图的距离
\setlength{\belowcaptionskip}{-0.6cm}%调整caption与下文的距离
\centering 
\includegraphics[width=17.4cm]{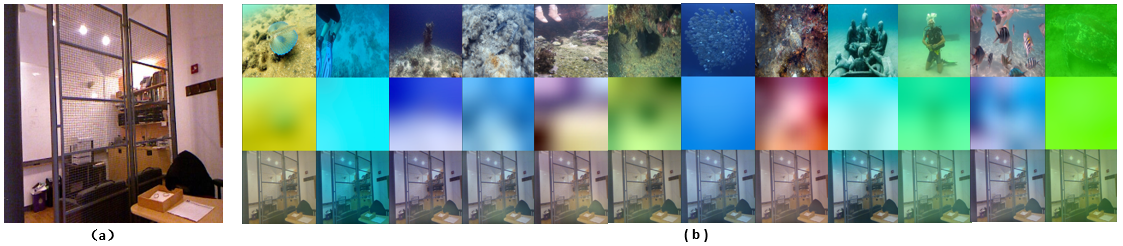}

\caption{Various synthetic underwater images are generated by the proposed UWNR network based on the same indoor clean image and different realistic underwater images. (a) The clean indoor image. (b) Realistic underwater images (top row), light field maps (middle row), and synthetic underwater images generated by the proposed method (bottom row).}
\label{fig:1}
\end{figure*}
For previous underwater image generation methods, commonly focus on physical model-based and GAN generating techniques \cite{Zhu_2017_ICCV,hou2020benchmarking,wang2021uwgan,wu2021novel}. It can be observed from Eq.(\ref{eq1}) that generating an underwater image requires estimating the ambient light $B(x)$ and the underwater transmission map $t(x)$. Moreover, physical model-based methods for underwater image generation still have the following limitations: (1) A simple physical model cannot cover all complex underwater scenes, especially the current physical model-based approach is biased by limited estimation of key parameters. (2) Previous methods most ignore the effect of the scattering coefficient in underwater imaging, because of the difficulty of capturing this coefficient accurately, which leads to unfavorable transmission maps
can cause the visual effect of generated underwater images to be un-satisfactory. (3) The underwater transmission map $t(x)$ of the physical model is a coefficient map with depth variation characteristics. Obtaining RGB-D images is relatively expensive and limited in resources.
%

% 这里之前写得不太好，我改了
To address these problems, we propose an underwater rendering framework to achieve the efficient rendering generation of true-to-life underwater images. The mechanism of \textbf{L}ight \textbf{F}ield \textbf{R}etention (LFR) in our framework can effectively transfer the diverse underwater style from natural underwater images to the objective generated images, which effectively guarantees the diversity of rendering results. It's worth to note that there is no physical model constraints and GAN-based method in our framework so that it can easily avoid following problems: (1) \textbf{Mode Collapse.} ~\cite{wang2021uwgan} (2) \textbf{Fake Features.}~\cite{wang2021uwgan} (3)\textbf{ Limited performance by incomplete underwater imaging physical model. \cite{hou2020benchmarking,liang2021single}} . And it is also worth mentioning that we utilize the well trained depth estimation network for collaborative work, which significantly alleviates the obstruction of our method in practical applications by avoiding the high cost of RGB-D images as the input of network~\cite{wang2021uwgan,li2017watergan}. 
Specifically, our method only needs a clean image and its depth map estimated by Li $et$ $al.$ \cite{li2018megadepth}'s way to generate a great quantity of underwater images from real world terrestrial image.
%.
To the best of our knowledge, this is the first method that gets rid of the physical model and GAN-based methods to generate underwater images.
We summarize our novelties and contributions as follows:
\begin{itemize}

\item We develop a natural light field retention module that renders the characteristics of the terrestrial image as close to the underwater situation as possible with the help of underwater dark channel loss and light field consistency loss we proposed.

\item To the best of our knowledge, this is the first work to render underwater image without physical models and GAN methods, which can easily render realistic underwater scenes with diverse styles.

\item We conduct extensive experiments to demonstrate that our method achieves state-of-the-art rendering performance in terms of objective evaluation.

\item We synthesize a large neural rendering underwater datasets (LNRUD), which contains a large number of underwater images synthesized from land images by our method.
\end{itemize}

\section{Related Work}
%\subsection{Underwater Image Enhancement}

{\bf Image Generation}. Generative adversarial networks were first proposed by Goodfellow $et$ $al.$ \cite{goodfellow2014generative}. In the field of image generation, many designs with adversarial theory as the base have produced unforgettable impression. CycleGAN \cite{Zhu_2017_ICCV} utilized cycle consistency principle to enhance GAN for unsupervised image style transfer, which also was extended to some other image generation methods and achieved amazing results \cite{press2020emerging,royer2020xgan,chang2018generating}. Motivated by the high quality of StyleGAN \cite{karras2019style,karras2020analyzing}, which performs exceptionally well on image editing and processing tasks. Based on the StyleGAN, Elad $et$ $al.$ \cite{eldred2021design} presented an end-to-end training method for learning a mapping from images to StyleGAN latent codes to generate high-quality images. Chong $et$ $al.$ \cite{chong2021stylegan} allowed StyleGAN to directly start various tasks through pre-training and a little operation on the latent space. In addition to these, there are many interesting image generation tasks using the GAN-based method\cite{li2021weather,sarkar2021humangan,chen2021intrinsic,marriott20213d}. For example, Li $et$ $al.$ \cite{li2021weather} developed Weather GAN to realize the transformation of different seasons and weather, Sarkar $et$ $al.$ \cite{sarkar2021humangan} designed HumanGAN which was the first way to address every aspect human image generation, e.g. global appearance sampling, pose transfer.

There are also inevitable problems behind the excellent effect of generative adversarial networks. The first is that the GAN architecture is usually difficult to train and it is not easy to objectively measure the generation effect \cite{arjovsky2017wasserstein,arjovsky2017principled,mescheder2018training,guo2020positive,ansari2020characteristic}. Another difficulty is that GAN will experience mode collapse due to improper punishment during training \cite{liu2019spectral,li2021tackling,srivastava2017veegan}.

{\bf Underwater Image Generation}. Deep learning has proven to perform well in various underwater tasks, but it is difficult to obtain large datasets in deep-sea environments. How to generate underwater images from existing resources is an important and challenging task \cite{li2017watergan,wang2021uwgan,hou2020benchmarking,anwar2018deep,wu2021novel,desai2021ruig}. In \cite{anwar2018deep,desai2021ruig}, authors constructed a large number of synthetic datasets utilizing previously calculated ocean attenuation coefficients combined with underwater attenuation models. Hou $et$ $al.$~\cite{hou2020benchmarking} developed the quadtree to select the background light area and obtained the transmission map according to the DCP principle~\cite{he2010single}. Different types of underwater degradation images were designed to imitate different underwater scenes. Li $et$ $al.$ \cite{li2017watergan} proposed WaterGAN, which takes in-air images, depth maps, and noise vectors as input and outputs synthetic images using a camera model and a generative adversarial network. On this basis, Wang $et$ $al.$ \cite{wang2021uwgan} presented UWGAN which only input aerial images and depth maps into the model, adopting the imaging model~\cite{ancuti2016dataset} and GAN to design a simple underwater image generation model.

The previous underwater image generation methods are based on the physical model as the vital core to render underwater images. Still, the biased estimation characteristics of the physical model make it difficult to guarantee the underwater image effect. The GAN-based approaches will lead to train unstably and prone to mode collapse. Our framework retains light field information to avoid the above problems and achieve better performance.

%###################### Method ############### 
\begin{figure*}[!t]
\vspace{-0.3cm}
\setlength{\abovecaptionskip}{0.1cm} %调整caption与图的距离
\setlength{\belowcaptionskip}{-0.6cm}%调整caption与下文的距离
    \centering
    \includegraphics[width=17.2cm]{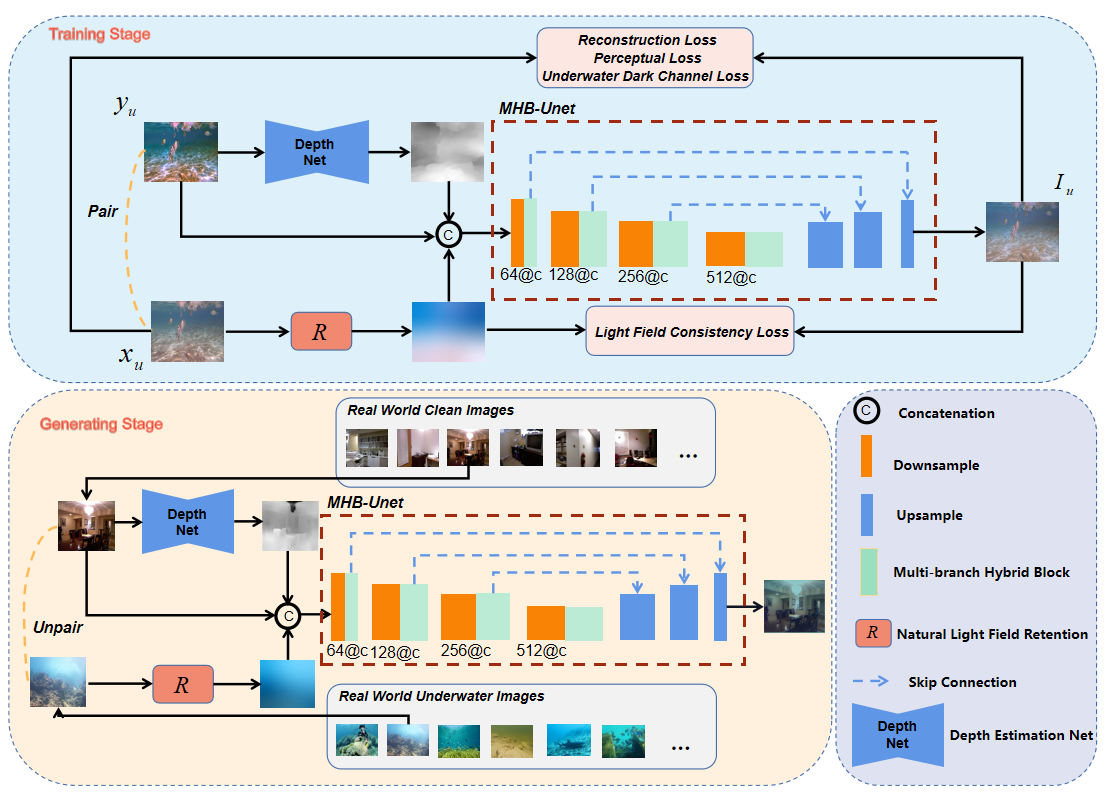}
    \caption{Underwater image rendering framework we proposed. In the training phase, with the pair of real underwater image and its clean ground-truth, the light field map  and  the scene depth are estimated using a natural light field retention module and a pre-trained depth estimation module respectfully. Then a MHB-Unet is trained to generate the synthetic underwater image. In the generating stage, a real underwater image can be used to render any unrelated clean image into an underwater image.}
    \label{fig:2}
\end{figure*}

\section{Proposed Method}
This section presents our underwater image rendering framework with natural light field retention. First, we discuss our natural light field retention module. Then we elaborate on the underwater image generation module of our method. Finally, we present the loss functions we used in training. The overview of our underwater rendering method is shown in Fig.~\ref{fig:2}.

\subsection{Natural Light Field Retention}
In theory, the transferring procedure consists of the estimation of three key parameters (\textit{i.e.}, $d(x)$, $\beta$ and $B(x)$) based on the underwater imaging model. Due to particle scattering, the light characteristics of underwater scenes are extremely different from terrestrial images because they have strong randomness, which causes difficulty for the traditional physical method to model these accurately. 

To address the above problem, we propose the light field retention scheme to transfer the diverse underwater light filed information to the objective image possibly. For an underwater image $x_{u}$ in dataset $\mathcal{X}_u\left\{ x_{u},y_{u}\right\}$, according to Retinex~\cite{land1977retinex,land1971lightness} theory, the image can be disassembled into 
\begin{equation}
    {x}_{u} = {x}_{l} \cdot {x}_{r}.
\end{equation}
where ${x}_{l}$ represents the illumination component of ambient light, ${x}_{r}$ is the reflection component of the target object carrying the image detail information. We perform a multi-scale Gaussian low-pass filter on the ${x}_{u}$ to obtain underwater light field map:
\begin{equation}
    {x}_{g} = \frac{1}{3}\sum_{\sigma}^{}Gauss_{\sigma}({x}_{u}),\sigma \in \left\{15,60,90 \right\}.
\end{equation}
similar to MSR~\cite{rahman1996multi}, we empirically set $\sigma$ to 15, 60, 90, respectively. The appropriate convolution kernel is selected by $\sigma$ adaptive control in the formula.
Considering that Gaussian filter may still contain object details, we transform it to logarithmic domain and scale it to get the final underwater light field map:
\begin{equation}
    {x}_{l} = \operatorname{Normalization}(\operatorname{log} \; {x}_{g}).
\end{equation}

Using the above operations, the natural light field maps of the real underwater images we obtained are shown in Fig.~\ref{fig:1}, from which we can found that our estimated light field does not contain object information and it well represents the underwater natural light field information. The reserved features in the underwater light field map focus on the natural style information of diverse underwater scenes without detailed and structured information from original underwater images. In theory, the underwater light field map contains two significant pieces of information for underwater characteristic transferring: $B(x)$ and $\beta$. It's worth noting that previous methods~\cite{wang2021uwgan,hou2020benchmarking} ignored the importance of $\beta$ for underwater imaging; recent work~\cite{li2021unsupervised} about hazy image generation calls our attention to focus on implicit estimation of the above two coefficients with entangled way. Our experiments demonstrate the obvious advantages of our transferring procedure.

\subsection{Underwater Image Generation Module}

{\bf Depth Estimation Network}. After the light field preservation module, we get the information of the light field. In addition, we also need to obtain the depth map that replaces the transmission map in the clean image image to preserve the depth information of the generated image. Therefore we design a depth estimation network whose network architecture we utilize Li $et$ $al.$'s pre-trained model \cite{li2018megadepth}. It is worth saying that our estimated depth map is a biased image. We feed this together into our underwater generative model (MHB-Unet) and let the network learn this biased property. The benefit is that our method does not require paired depth map datasets and can synthesize a large number of underwater images for practicality.

\begin{figure}[!h]
\vspace{-0.2cm}
    \centering
    \includegraphics[width=8cm]{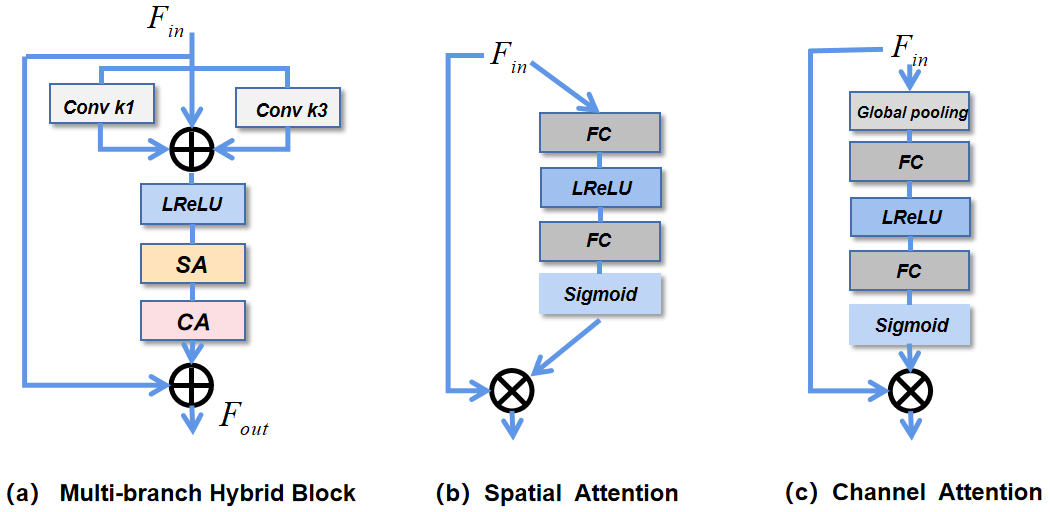}
    \caption{From left to right are multi-branch hybrid modules, spatial attention and channel attention, respectively.}
    \label{fig:6}
\end{figure}

{\bf Multi-branch Hybrid Unet (MHB-Unet)}. We design the Unet style architecture network as our underwater image generation model, called MHB-Unet. For further improving the performance of MHB-Unet, we develop a multi-scale hybrid convolutional attention module as shown in Fig.~\ref{fig:6}(a). Considering that the local features of underwater scenes are complex and diverse, we first obtain different receptive fields through 1×1 and 3×3 convolutions to perform multiple feature fusion. At the same time, we also use residual connections, which can solve the problem of vanishing gradients and take into account that the spatial structure and color of certain regions in underwater images are not affected by scene degradation. After multi-branch fusion, we apply a combination of spatial attention Fig.~\ref{fig:6}(b) and channel attention module Fig.~\ref{fig:6}(c). The spatial attention mechanism improves the network's ability to pay attention to complex areas such as light field distribution and depth information in underwater images, while the channel attention pays attention to the network's expression of important channels in features, thereby improving the overall expression performance of the model.

It is worth noting that our model has excellent performance and even high-resolution images can be rendered well. Please refer to the more discussion section of experiments for more details of actual performance.
% We adopt Unet as our backbone, whose upsampling and downsampling processes extract shallow and deep features respectively. We add our multi-scale hybrid block after each downsampling to reduce the loss of information and utilize the multi-scale information to enhance the information exchange of skip connections.

\subsection{Training Losses}
We propose following loss functions to ensure the rendering effect of the generated underwater images.

{\bf Reconstruction Loss}. To ensure that our predicted images are enough close to the real underwater images, we introduce the $L_1$ loss as our basic reconstruction loss:
\begin{equation}
    \mathcal{L}_{rec} = \left\|\mathcal{I}_{u}-x_{u}\right\|_{1}.
\end{equation}
where $\left\| \cdot \right\|$ represents L1 loss, $\mathcal{I}_{u}$ is the output underwater image by our architecture.

{\bf Perceptual Loss}. Perceptual loss which utilizes the feature layers extracted from pre-trained VGG19 model \cite{simonyan2014very} as the loss network aims to maintain perceptual structure consistency. It is defined as follows:
\begin{equation}
    \mathcal{L}_{\mathrm{per}}=\sum_{j=1}^{5} \frac{1}{C_{j} H_{j} W_{j}}\left\|\phi_{j}\left(\mathcal{I}_{u}\right)-\phi_{j}\left(x_{u})\right)\right\|_{2}^{2}.
\end{equation}
where $C_{j}$ $H_{j}$ $W_{j}$ represents the number of channels, length and width of the feature map extracted in the j-th hidden features respectively. $\phi_{j}$ is the specified j-th layer of the loss network.

{\bf Underwater Dark Channel Loss}. UDCP \cite{drews2013transmission} applied the dark channel prior principle to underwater. We define an underwater dark channel loss to make the generated underwater image consistent with the clean image at the dark channel level.
\begin{equation}
   U D C\left(x\right)=\min _{y \in N(x)}\left[\min _{c \in\{g, b\}} {x_{i}}^{c}(y)\right].
\end{equation}
where $x$ and $y$ are pixel coordinates of image $x_{i}$, $x_{c}$ represents c-th color channel of $x_{i}$, and N(x) denotes the local neighborhood centered at $x_{i}$. 

The formula for the underwater dark channel loss is as follows:
\begin{equation}
    \mathcal{L}_{udc}=\left\|UDC\left(\mathcal{I}_{u}\right)-UDC\left(x_{u}\right)\right\|_{1}.
\end{equation}

{\bf Light Field Consistency Loss}. In order to effectively maintain the light field characteristic of real underwater images, we introduce the light field consistency loss based on the natural light field map for better rendering performance. We utilize multi-scale Gaussian filter to capture the light field map: 
\begin{equation}
    \mathcal{LF}(\mathcal{J})= \frac{1}{3}\sum_{\sigma}^{}Gauss_{\sigma}(\mathcal{J}),\sigma \in \left\{15,60,90 \right\}.
\end{equation}
where the $\mathcal{J}$ denotes the image and $\mathcal{LF}(\cdot)$ denotes the capturing operation of light filed. And the light field consistency loss function is defined as follows: 
\begin{equation}
    L_{lfc} = \left\|\mathcal{LF}(\mathcal{I}_{u})-\mathcal{LF}(x_{l}) \right\|_{1}.
\end{equation}

{\bf Overall Loss Function}. The overall loss function is expressed as follows:
\begin{equation}
    \mathcal{L} = \lambda_{rec}\mathcal{L}_{rec}+ \lambda_{per}\mathcal{L}_{per}+ \lambda_{udc}\mathcal{L}_{udc}+ \lambda_{lfc}\mathcal{L}_{lfc}.
\end{equation}
where $\lambda_{rec}$, $\lambda_{per}$, $\lambda_{udc}$ and $\lambda_{lfc}$ are trade-off weights.

\section{Experiments}

In this section we conduct extensive experiments to evaluate the effectiveness of our method. First we describe the specific implementation details of our training part. Second, we adopt the FID~\cite{heusel2017gans} evaluation metric to objectively evaluate the effect of our generated images, then we adopt PSNR, SSIM and UIQM metrics to measure the effect of our generated underwater image dataset compared with other underwater image generation methods on the underwater enhancement network. Finally we perform ablation experiments to vertify the necessity of the components in our framework.

\subsection{Experiment Settings}
% %
% \begin{table*}[!t]
% \centering

% \caption{Comparative metric analysis of various methods on NYU datasets} \label{table:3} %表标题
% \begin{tabular}{ccccc}
% \hline
%  & FID & PSNR & SSIM & UIQM \\ \hline
% Clean images &  & - & - & -\\ 
% UWGAN &  &  &  \\ 
% Ours(UWNR) &   &  & \\ \hline
% \end{tabular}}
% \end{table*}

% \begin{table*}[!t]
% \caption{Comparative metric analysis of various methods on SUID datasets} \label{table:3} %表标题
% \centering
% \setlength{\tabcolsep}{6mm}{

% \begin{tabular}{ccccc}
% \toprule

%  & FID & PSNR & SSIM & UIQM \\ \hline
% Clean images &  & - & - & - \\ 
% UISA &  &  &  \\ 
% Ours(UWNR) &   &  & \\ 
% \bottomrule
% \end{tabular}}
% \end{table*}

\begin{table*}[!t]
\centering
\caption{Comparative metric analysis of various methods on different datasets. Best results marked in bold.} \label{table:Comaprativemetric}
\resizebox{16cm}{2.0cm}{
\renewcommand\arraystretch{1.5}
\begin{tabular}{p{2cm}<{\centering}|p{2cm}<{\centering}p{2cm}<{\centering}p{2cm}<{\centering}|p{2cm}<{\centering}p{2cm}<{\centering}p{2cm}<{\centering}}
\toprule[1.2pt]
%\specialrule{1em}{1pt}{1pt}
\multirow{2}{*}{Metrics} & \multicolumn{3}{c|}{NYU Dataset(890)} & \multicolumn{3}{c}{SUID Dataset(890)} \\ \cline{2-7} 
 & \multicolumn{1}{c|}{Clean Images} & \multicolumn{1}{c|}{UWGAN \cite{wang2021uwgan}} & \multicolumn{1}{c|}{Ours (UWNR)} & \multicolumn{1}{c|}{Clean Images} & \multicolumn{1}{c|}{UISA \cite{hou2020benchmarking}} & Ours(UWNR) \\ \hline 
FID & 239.36 & 236.23 & $\mathbf{221.93}$ & 274.67 & 220.90 & $\mathbf{216.76}$ \\  
PSNR & - & 18.41 & $\mathbf{18.86}$ & - & 12.87 &  $\mathbf{19.32}$\\ 
SSIM & - & 0.70 & $\mathbf{0.77}$ & - & 0.63 &  $\mathbf{0.77}$\\ 
UIQM & - & 2.28 & $\mathbf{2.62}$ & - & 2.44 &  $\mathbf{2.63}$\\ 
\bottomrule [1.2pt]
\end{tabular}}
\end{table*}
\begin{figure*}[!t]
\vspace{-0.3cm}
\setlength{\abovecaptionskip}{0.1cm} %调整caption与图的距离
\setlength{\belowcaptionskip}{-0.4cm}%调整caption与下文的距离
\centering 
\includegraphics[width=14cm]{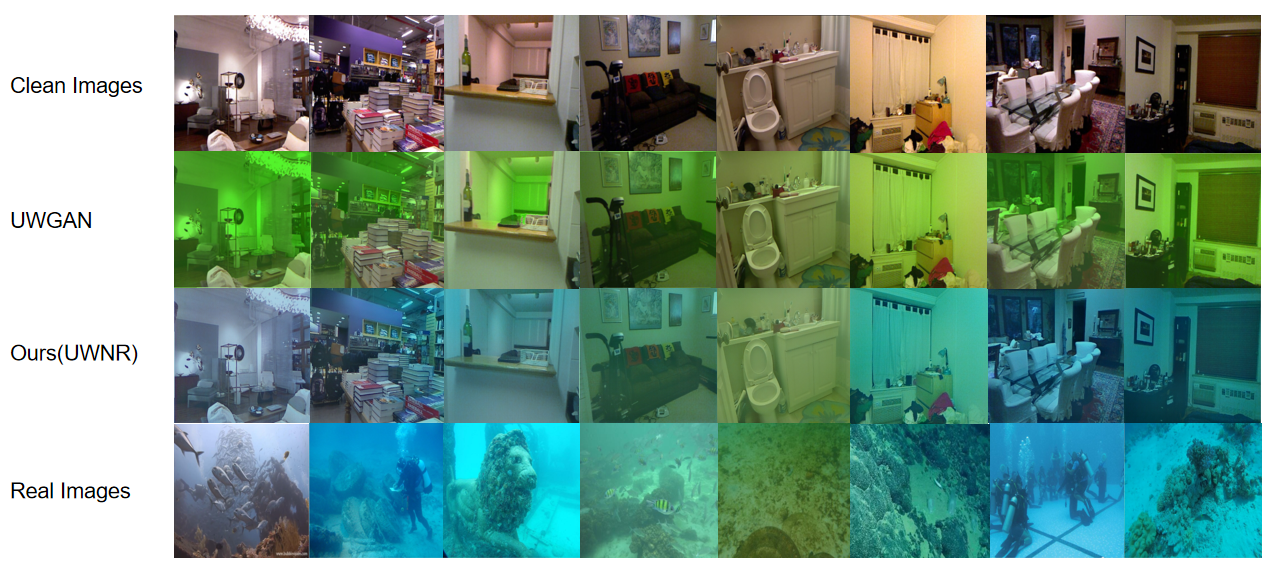}
\caption{Underwater images synthesized by UWGAN\cite{wang2021uwgan} and our method (UWNR) on the NYU\cite{Silberman:ECCV12} dataset.}\label{fig:4}
\end{figure*}
\begin{figure*}[!t]

\centering 
%\vspace{-0.3cm}
\setlength{\abovecaptionskip}{0.1cm} %调整caption与图的距离
\setlength{\belowcaptionskip}{-0.4cm}%调整caption与下文的距离
\includegraphics[width=14cm]{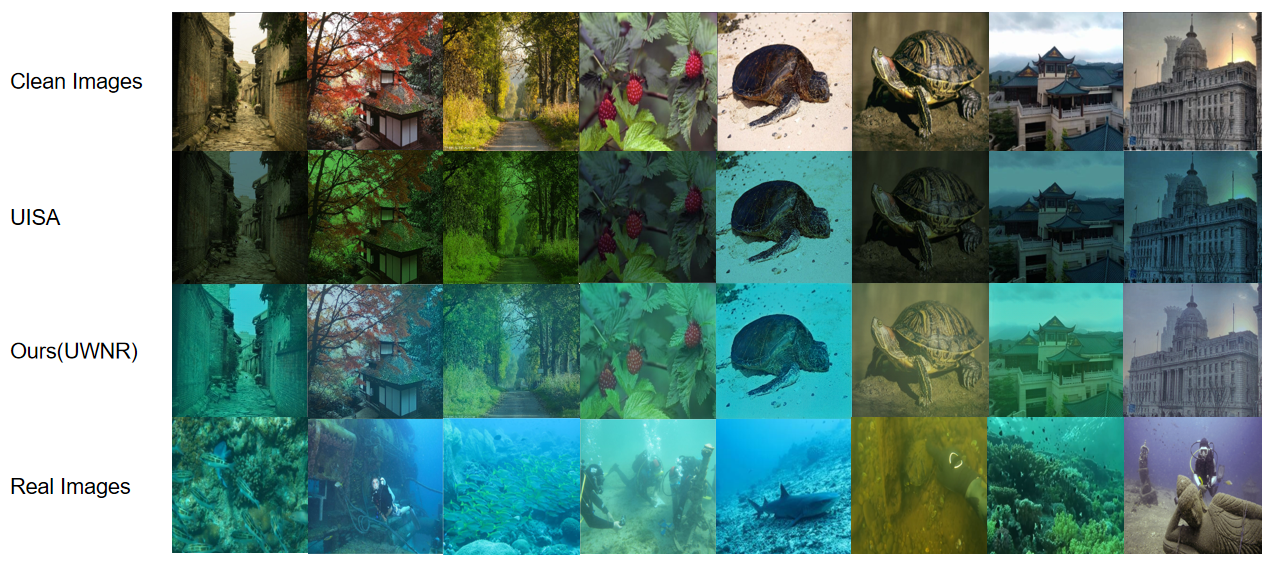}
\caption{Underwater images synthesized by UISA\cite{hou2020benchmarking} and our method (UWNR) on the SUID\cite{hou2020benchmarking} dataset.}\label{fig:4.5}
\end{figure*}
{\bf Training Details}. In training details, we select 500 images from the underwater real dataset UIEB \cite{li2019underwater} as our training set. We also add random rotations between 0, 90, 180 and 270 degrees and horizontal flip for data augmentation. The patch of the input for training is set to 256×256 and we train the network for 200 epochs, where we set the initial learning rate to 0.0002, and after 100 epochs we start a linear decay of the learning rate. We employ the Adam optimizer whose first momentum and second momentum are taken to be 0.9 and 0.999, respectively. For perceptual loss, we select layers 1, 3, 5, 9 and 13 in the VGG19~\cite{simonyan2014very} model to extract hidden features.

\begin{table*}[!t]

\vspace{-0.3cm}
\setlength{\abovecaptionskip}{0.1cm} %调整caption与图的距离
\setlength{\belowcaptionskip}{-1.4cm}%调整caption与下文的距离

\centering
\caption{Ablation study on different loss function configurations of our UWNR framework, bold means the best result.}\label{tab:ablation_loss}
\resizebox{17cm}{2.2cm}{
\renewcommand\arraystretch{1.5}
\begin{tabular}{p{2.1cm}<{\centering}|p{2.1cm}<{\centering}p{1.5cm}<{\centering}p{1.5cm}<{\centering}p{1.5cm}<{\centering}|p{1.5cm}<{\centering}p{1.5cm}<{\centering}p{1.5cm}<{\centering}p{1.5cm}<{\centering}}
\toprule[1.2pt] \multirow{2}{*}{ Method } & \multicolumn{4}{c|}{ NYU Dataset(890) } & \multicolumn{4}{c}{ SUID Dataset(890) } \\
\cline { 2 - 9 } & FID & PSNR & SSIM & UIQM & FID & PSNR & SSIM & UIQM \\
\hline \hline Ours w/o $L_{rec}$ & $224.97$ & $18.01$ & $0.76$ & $2.54$ & $224.68$ & $18.22$ & 0.76 & 2.55\\
\hline Ours w/o $L_{\text {per}}$ & $\mathbf{216.42}$ & $17.55$ & $0.67$ & $2.55$ & $223.54$ & $18.48$ & 0.70& 2.59\\
\hline Ours w/o $L_{\text {udc}}$ & $223.09$ & $18.24$ & ${0.77}$ & $2.59$ & $218.10$ & $18.89$ & 0.76& 2.61\\
\hline Ours w/o $L_{\text {lfc}}$ & $223.81$ & $18.21$ & $0.76$ & $2.60$ & $220.48$ & $18.73$ &0.76 & 2.63\\
\hline Ours & ${221.93}$ & $\mathbf{18.86}$ & $\mathbf{0.77}$ & $\mathbf{2.62}$ & $\mathbf{216.76}$ & $\mathbf{19.32}$ & $\mathbf{0.77}$& $\mathbf{2.64}$\\\bottomrule[1.2pt]
    \end{tabular}}

\end{table*}

{\bf Evaluation Metrics}. How to evaluate the data generated by the image generation network is a vital question. The generated dataset should be visually appealing, diverse, and close to the real domain. Based on the above, we choose FID~\cite{heusel2017gans} as the evaluation metrics. FID directly considers the distance between the synthetic data and the real data at the feature level to measure the difference between the generated image and the real image. The smaller the value, the closer it is to the image in the real domain.
To measure the performance of our synthetic dataset, we use it with other synthetic underwater datasets to test the recovered results in the underwater image enhancement network. We apply the Peak Signal-to-Noise Ration (PSNR) and Structural Similarity (SSIM) to objectively evaluate the enhanced performance. In addition, we also use the non-reference indicator UIQM, which evaluates colorfulness (UICM), sharpness (UISM), and contrast (UIConM).

\subsection{Quantitative Comparison}
We apply the various real underwater image $x_{r}$ in $\mathcal{X}_r$ $\left\{x_{r}, y_{r}\right\}$ as the reference for the FID evaluation metric, and its distance from the generated underwater dataset measures the reality of the data. We compare our method with previous state-of-the-art underwater image synthesis methods. We select another part of the light field map that is different from the training details to generate images so as to avoid the risk of data leakage. Since the two methods adopt NYU \cite{Silberman:ECCV12} and SUID \cite{hou2020benchmarking} datasets to generate underwater datasets. For the sake of fairness using FID metric, we compare our method with previous two rendering methods on their datasets respectively. Specifically, for a fair and efficient comparison, we randomly select 890 clean images from the NYU and SUID datasets, using UWGAN\cite{wang2021uwgan} and UISA\cite{hou2020benchmarking} methods to generate corresponding underwater images, and we render the clean images into the underwater images by our method, which also are included in our LNRU dataset.
In the visual comparison, as shown in Tab.~\ref{table:Comaprativemetric}, we can observe that the results of our method are closer to the natural underwater images than other methods.

At the same time, in order to verify the effectiveness of our method, we use our dataset and the datasets of other methods where the contents of the scenes are the same to train the underwater enhancement network Shallow-UWnet~\cite{naik2021shallow}, in which we train for 100 epochs, the other training details are the same as those in the original paper. For the test part, we adopt 90 small image dataset in the UIEB dataset to measure the effect of underwater enhancement by utilizing PSNR,SSIM and UIQM metrics.

%\vspace{-0.8cm}
~\paragraph{Model Complexity Analysis.}
We analyze the parameters of our model is 11.57M. We also study the computation and inferencing runtime of rendering is 276.26GMac/0.0023s when the image size is 1024$\times$1024, which illustrates the efficiency of our model during generating stage and provides a prerequisite for the actual project landing.

\subsection{Visual Comparison}
To verify the superiority and diversity of our proposed method, we compare the  state-of-the-art method of underwater image synthesis and present the results in Fig.~\ref{fig:4} and Fig.~\ref{fig:4.5}. Due to the mode collapse issue of the GAN architecture, the UWGAN \cite{wang2021uwgan} method in the NYU dataset results in a single underwater dataset which only has the green underwater feature. Compared with UWGAN, our method can generate more diverse underwater images with better visual effects.

In the SUID dataset, the UISA~\cite{hou2020benchmarking} method based physical model estimation to generate different types of underwater images. But the estimated deviation of its parameters will cause a large visual difference between the synthetic underwater image and the real underwater environment, which makes the generated underwater images susceptible to appear too bright or too dark to match the real underwater scene. Our framework does not have this problem in terms of visual effects and is more in line with underwater characteristics.

\subsection{Ablation Study}
To validate the necessity of the components in our approach, we performed extensive ablation experiments to validate our claims.

As shown in Tab.~\ref{tab:ablation1}, we perform ablation experiments of the light field retention (LFR) module and the depth estimation network (DEN) to show their necessity. We eliminate these two modules separately to carry out the following training: (1) LFR+MHB-Unet. (2) DEN+DCP+MHB-Unet. To demonstrate the superiority of our light field preserving module, we replace our light field module with a dark channel prior to obtain background light.(3) LFR+DEN+MHB-Unet (Ours).
\begin{table}[!h]
%\vspace{-0.5cm}
\setlength{\belowcaptionskip}{0.2cm}%调整caption与下文的距离
\setlength{\abovecaptionskip}{0.3cm} %调整caption与图的距离

\centering
\caption{Ablation experiments with light field retention and depth estimation modules. Bold presents best result.}\label{tab:ablation2}
\resizebox{8.3cm}{1.25cm}{
\renewcommand\arraystretch{1.5}
\begin{tabular}{c|cccc|cccc}
\toprule[1.2pt] \multirow{2}{*}{ Method } & \multicolumn{4}{c|}{ NYU Dataset(890) } & \multicolumn{4}{c}{ SUID Dataset(890) } \\
\cline { 2 - 9 } & FID & PSNR & SSIM & UIQM & FID & PSNR & SSIM & UIQM \\
\hline  Ours w/o LFR+DCP & $223.52$ & $18.58$ & $0.76$ & $2.61$ & $232.47$ & $17.64$ & 0.71 & 2.64\\
\hline Ours w/o DEN & $223.36$ & $18.45$ & $0.76$ & $2.57$ & $221.96$ & $18.82$ & 0.73 & 2.59\\
\hline Ours & $\mathbf{221.93}$ & $\mathbf{18.86}$ & $\mathbf{0.77}$ & $\mathbf{2.62}$ & $\mathbf{216.76}$ & $\mathbf{19.32}$ & $\mathbf{0.77}$& $\mathbf{2.64}$\\\bottomrule[1.2pt]
    \end{tabular}}
\end{table}
From the Tab.~\ref{tab:ablation2}, we can observe that the lack of depth information and light field information will reduce the quantitative performance of our method. 
In particular, as shown in Fig.~\ref{fig:8}, compared to the background light obtained by DCP~\cite{he2010single}, our light field retention method can reduce the limitation of diversity and generate a more progressive effect.
\begin{figure}[!h]
\vspace{-0.2cm}
\setlength{\abovecaptionskip}{0.1cm} %调整caption与图的距离
\setlength{\belowcaptionskip}{-0.5cm}%调整caption与下文的距离
\centering 
\includegraphics[width=8cm]{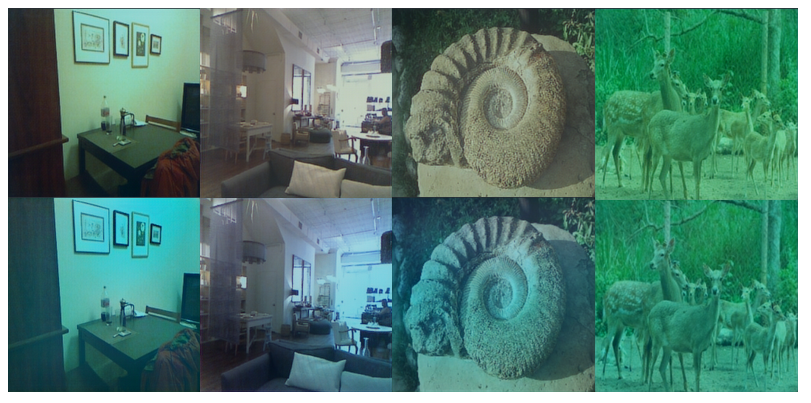}
\caption{Top is the underwater image which applys DCP\cite{he2010single} method to replace our LFR module, bottom is the underwater image with LFR.}\label{fig:8}
\end{figure}

\begin{figure}[!h]
%\vspace{-0.5cm}
\setlength{\abovecaptionskip}{0.1cm} %调整caption与图的距离
\setlength{\belowcaptionskip}{-0.5cm}%调整caption与下文的距离
\centering 
\includegraphics[width=8cm]{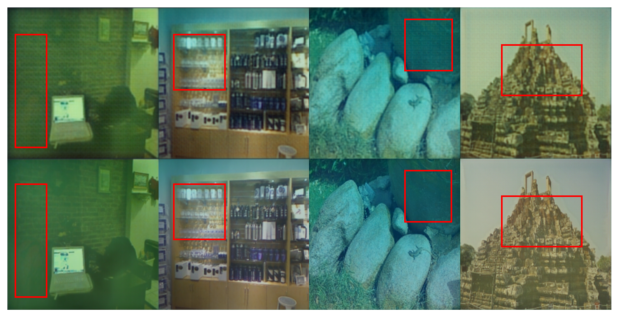}
\caption{Top is the underwater image w/o perceptual loss, bottom is the underwater image with perceptual loss.}\label{fig:5}
\end{figure}

To elaborate the effectiveness of our loss function, we also perform extensive ablation experiments to eliminate each loss function separately to train our model and test it. From the Tab.~\ref{tab:ablation_loss}, we observe that the absence of reconstruction loss causes a significant drop in metrics because it is critical for pixel-level reconstruction of the image. The loss of underwater dark channel is considered from the statistical characteristics of underwater images, and the light field consistency loss is based on the characteristics of underwater ambient light, which provide conditions for underwater image rendering from different angles. Not using them can also cause model performance to drop.
As show in Fig.~\ref{fig:5}, it is worth noting that removing the perceptual loss will make the indicator FID of underwater synthesis more excellent. But in terms of intuitive visual effects, we will find that the loss of detail information is serious, especially in some contours, which is a fatal issue in image synthesis.

And we perform another ablation experiment to demonstrate the effectiveness of the multi-branch Hybrid Block in our MHB-Unet. We present the following three training methods: (1) {Unet+Multi-Branch Hybrid Block w/o spatial attention}. (2) {Unet+Multi-Branch Hybrid Block w/o channel attention}. (3) {Unet+Multi-Branch Hybrid Block w/o Multi-branch Hybrid Convolution (MHC)}. From the results in Tab.~\ref{tab:ablation1}, we prove that the multi-branch-based channel attention and spatial attention can better allow our model to learn the characteristics of real underwater complex environments and achieve better metrics. We also found that the absence of multi-branch convolutions and residual connections reduces our metric performance. Probably because of the lack of multiple feature fusion, the processing power in diverse and complex regions is reduced.
\begin{table}[!h]
\centering
\caption{The network architecture ablation experiment of MHB-Unet, bold indicates the best result.}\label{tab:ablation1}
\resizebox{8.3cm}{1.5cm}{
\renewcommand\arraystretch{1.5}
\begin{tabular}{c|cccc|cccc}
\toprule[1.2pt] \multirow{2}{*}{ Method } & \multicolumn{4}{c|}{ NYU Dataset(890) } & \multicolumn{4}{c}{ SUID Dataset(890) } \\
\cline { 2 - 9 } & FID & PSNR & SSIM & UIQM & FID & PSNR & SSIM & UIQM \\
\hline \hline Ours w/o SA & $223.55$ & $18.05$ & $0.75$ & $2.57$ & $218.21$ & $18.48$ & 0.76 & 2.57\\
\hline Ours w/o CA & ${224.69}$ & $18.18$ & $0.76$ & $2.58$ & $222.60$ & $18.67$ & 0.77& 2.58\\
\hline Ours w/o MHC & $222.2$ & $18.71$ & $0.75$ & $2.61$ & $220.61$ & $19.10$ &0.74 & 2.63\\
\hline Ours &$\mathbf{221.93}$ & $\mathbf{18.86}$ & $\mathbf{0.77}$ & $\mathbf{2.62}$ & $\mathbf{216.76}$ & $\mathbf{19.32}$ & $\mathbf{0.77}$& $\mathbf{2.64}$\\\bottomrule[1.2pt]
    \end{tabular}}

\end{table}
\vspace{-0.5cm}

\subsection{Large Neural Rendering Underwater (LNRU) Dataset}
We create a large neural rendering underwater dataset with 50,000 underwater images, in which each underwater image has the associate groundtruth of a latent clean image. Specifically, we collected 5,000 authentic underwater images and randomly selected them for rendering. We will open-source the LNRU dataset after our submitted paper is received. Please refer to our supplementary material for reviewing the thumbnails and more details of LNRU dataset.
\section{Conclusion}
In this paper, we propose an underwater image rendering framework that avoids the problem of ground truth inaccuracy in underwater paired datasets. We utilize light field information and a multi-scale unet network to generate a large number of diverse images using only unpaired images during rendering. In addition to this, we have open-sourced a large-scale underwater image synthesis dataset. The experimental results demonstrate that our method achieves the best results compared with the state-of-the-art methods in terms of vision and metrics.
\vspace{-0.5cm}
% \paragraph{\textit{Limitations:}}Our neural rendering network still has to learn the mapping from paired underwater datasets, which limits the performance of our method. We hope to further explore the neural rendering by unsupervised learning in the future. 
{\small
\bibliographystyle{ieee_fullname}
\bibliography{egbib}

\begin{thebibliography}{10}\itemsep=-1pt

\bibitem{ancuti2016dataset}
C Ancuti, CO Ancuti, CD Vleeschouwer, et~al.
\newblock A dataset to evaluate quantitatively dehazing algorithms.
\newblock In {\em International Conference on Image Processing (ICIP 2016)}.

\bibitem{ansari2020characteristic}
Abdul~Fatir Ansari, Jonathan Scarlett, and Harold Soh.
\newblock A characteristic function approach to deep implicit generative
  modeling.
\newblock In {\em Proceedings of the IEEE/CVF Conference on Computer Vision and
  Pattern Recognition}, pages 7478--7487, 2020.

\bibitem{anwar2018deep}
Saeed Anwar, Chongyi Li, and Fatih Porikli.
\newblock Deep underwater image enhancement.
\newblock {\em arXiv preprint arXiv:1807.03528}, 2018.

\bibitem{arjovsky2017principled}
Martin Arjovsky and Léon Bottou.
\newblock Towards principled methods for training generative adversarial
  networks, 2017.

\bibitem{arjovsky2017wasserstein}
Martin Arjovsky, Soumith Chintala, and Léon Bottou.
\newblock Wasserstein gan, 2017.

\bibitem{bailey2008archaeology}
Geoffrey~N Bailey and Nicholas~C Flemming.
\newblock Archaeology of the continental shelf: marine resources, submerged
  landscapes and underwater archaeology.
\newblock {\em Quaternary Science Reviews}, 27(23-24):2153--2165, 2008.

\bibitem{chang2018generating}
Bo Chang, Qiong Zhang, Shenyi Pan, and Lili Meng.
\newblock Generating handwritten chinese characters using cyclegan.
\newblock In {\em 2018 IEEE winter conference on applications of computer
  vision (WACV)}, pages 199--207. IEEE, 2018.

\bibitem{chen2021intrinsic}
Haoyu Chen, Hao Tang, Henglin Shi, Wei Peng, Nicu Sebe, and Guoying Zhao.
\newblock Intrinsic-extrinsic preserved gans for unsupervised 3d pose transfer.
\newblock In {\em Proceedings of the IEEE/CVF International Conference on
  Computer Vision}, pages 8630--8639, 2021.

\bibitem{chong2021stylegan}
Min~Jin Chong, Hsin-Ying Lee, and David Forsyth.
\newblock Stylegan of all trades: Image manipulation with only pretrained
  stylegan.
\newblock {\em arXiv preprint arXiv:2111.01619}, 2021.

\bibitem{desai2021ruig}
Chaitra Desai, Ramesh~Ashok Tabib, Sai~Sudheer Reddy, Ujwala Patil, and Uma
  Mudenagudi.
\newblock Ruig: Realistic underwater image generation towards restoration.
\newblock In {\em Proceedings of the IEEE/CVF Conference on Computer Vision and
  Pattern Recognition}, pages 2181--2189, 2021.

\bibitem{drews2013transmission}
Paul Drews, Erickson Nascimento, Filipe Moraes, Silvia Botelho, and Mario
  Campos.
\newblock Transmission estimation in underwater single images.
\newblock In {\em Proceedings of the IEEE international conference on computer
  vision workshops}, pages 825--830, 2013.

\bibitem{eldred2021design}
Ross Eldred, Johnathan Lussier, and Anthony Pollman.
\newblock Design and testing of a spherical autonomous underwater vehicle for
  shipwreck interior exploration.
\newblock {\em Journal of Marine Science and Engineering}, 9(3):320, 2021.

\bibitem{goodfellow2014generative}
Ian Goodfellow, Jean Pouget-Abadie, Mehdi Mirza, Bing Xu, David Warde-Farley,
  Sherjil Ozair, Aaron Courville, and Yoshua Bengio.
\newblock Generative adversarial nets.
\newblock {\em Advances in neural information processing systems}, 27, 2014.

\bibitem{gu2018overview}
Linyi GU, Qi SONG, Hongwei YIN, and Jie JIA.
\newblock An overview of the underwater search and salvage process based on
  rov.
\newblock {\em SCIENTIA SINICA Informationis}, 48(9):1137--1151, 2018.

\bibitem{guo2020positive}
Tianyu Guo, Chang Xu, Jiajun Huang, Yunhe Wang, Boxin Shi, Chao Xu, and Dacheng
  Tao.
\newblock On positive-unlabeled classification in gan.
\newblock In {\em Proceedings of the IEEE/CVF Conference on Computer Vision and
  Pattern Recognition}, pages 8385--8393, 2020.

\bibitem{he2010single}
Kaiming He, Jian Sun, and Xiaoou Tang.
\newblock Single image haze removal using dark channel prior.
\newblock {\em IEEE transactions on pattern analysis and machine intelligence},
  33(12):2341--2353, 2010.

\bibitem{heusel2017gans}
Martin Heusel, Hubert Ramsauer, Thomas Unterthiner, Bernhard Nessler, and Sepp
  Hochreiter.
\newblock Gans trained by a two time-scale update rule converge to a local nash
  equilibrium.
\newblock {\em Advances in neural information processing systems}, 30, 2017.

\bibitem{hou2020benchmarking}
Guojia Hou, Xin Zhao, Zhenkuan Pan, Huan Yang, Lu Tan, and Jingming Li.
\newblock Benchmarking underwater image enhancement and restoration, and
  beyond.
\newblock {\em IEEE Access}, 8:122078--122091, 2020.

\bibitem{islam2020simultaneous}
Md~Jahidul Islam, Peigen Luo, and Junaed Sattar.
\newblock Simultaneous enhancement and super-resolution of underwater imagery
  for improved visual perception, 2020.

\bibitem{islam2020fast}
Md~Jahidul Islam, Youya Xia, and Junaed Sattar.
\newblock Fast underwater image enhancement for improved visual perception.
\newblock {\em IEEE Robotics and Automation Letters}, 5(2):3227--3234, 2020.

\bibitem{jahanbakht2021internet}
Mohammad Jahanbakht, Wei Xiang, Lajos Hanzo, and Mostafa~Rahimi Azghadi.
\newblock Internet of underwater things and big marine data analytics—a
  comprehensive survey.
\newblock {\em IEEE Communications Surveys \& Tutorials}, 2021.

\bibitem{karras2019style}
Tero Karras, Samuli Laine, and Timo Aila.
\newblock A style-based generator architecture for generative adversarial
  networks.
\newblock In {\em Proceedings of the IEEE/CVF conference on computer vision and
  pattern recognition}, pages 4401--4410, 2019.

\bibitem{karras2020analyzing}
Tero Karras, Samuli Laine, Miika Aittala, Janne Hellsten, Jaakko Lehtinen, and
  Timo Aila.
\newblock Analyzing and improving the image quality of stylegan.
\newblock In {\em Proceedings of the IEEE/CVF conference on computer vision and
  pattern recognition}, pages 8110--8119, 2020.

\bibitem{komatsu2018optical}
Satoru Komatsu, Adam Markman, and Bahram Javidi.
\newblock Optical sensing and detection in turbid water using multidimensional
  integral imaging.
\newblock {\em Optics letters}, 43(14):3261--3264, 2018.

\bibitem{land1977retinex}
Edwin~H Land.
\newblock The retinex theory of color vision.
\newblock {\em Scientific american}, 237(6):108--129, 1977.

\bibitem{land1971lightness}
Edwin~H Land and John~J McCann.
\newblock Lightness and retinex theory.
\newblock {\em Josa}, 61(1):1--11, 1971.

\bibitem{li2021unsupervised}
Boyun Li, Yijie Lin, Xiao Liu, Peng Hu, Jiancheng Lv, and Xi Peng.
\newblock Unsupervised neural rendering for image hazing, 2021.

\bibitem{li2019benchmarking}
Boyi Li, Wenqi Ren, Dengpan Fu, Dacheng Tao, Dan Feng, Wenjun Zeng, and
  Zhangyang Wang.
\newblock Benchmarking single-image dehazing and beyond.
\newblock {\em IEEE Transactions on Image Processing}, 28(1):492--505, 2019.

\bibitem{li2019underwater}
Chongyi Li, Chunle Guo, Wenqi Ren, Runmin Cong, Junhui Hou, Sam Kwong, and
  Dacheng Tao.
\newblock An underwater image enhancement benchmark dataset and beyond, 2019.

\bibitem{li2017watergan}
Jie Li, Katherine~A Skinner, Ryan~M Eustice, and Matthew Johnson-Roberson.
\newblock Watergan: Unsupervised generative network to enable real-time color
  correction of monocular underwater images.
\newblock {\em IEEE Robotics and Automation letters}, 3(1):387--394, 2017.

\bibitem{li2021tackling}
Wei Li, Li Fan, Zhenyu Wang, Chao Ma, and Xiaohui Cui.
\newblock Tackling mode collapse in multi-generator gans with orthogonal
  vectors.
\newblock {\em Pattern Recognition}, 110:107646, 2021.

\bibitem{li2021weather}
Xuelong Li, Kai Kou, and Bin Zhao.
\newblock Weather gan: Multi-domain weather translation using generative
  adversarial networks.
\newblock {\em arXiv preprint arXiv:2103.05422}, 2021.

\bibitem{li2018megadepth}
Zhengqi Li and Noah Snavely.
\newblock Megadepth: Learning single-view depth prediction from internet
  photos.
\newblock In {\em Proceedings of the IEEE Conference on Computer Vision and
  Pattern Recognition}, pages 2041--2050, 2018.

\bibitem{liang2021single}
Zheng Liang, Yafei Wang, Xueyan Ding, Zetian Mi, and Xianping Fu.
\newblock Single underwater image enhancement by attenuation map guided color
  correction and detail preserved dehazing.
\newblock {\em Neurocomputing}, 425:160--172, 2021.

\bibitem{liu2019spectral}
Kanglin Liu, Wenming Tang, Fei Zhou, and Guoping Qiu.
\newblock Spectral regularization for combating mode collapse in gans.
\newblock In {\em Proceedings of the IEEE/CVF International Conference on
  Computer Vision}, pages 6382--6390, 2019.

\bibitem{marriott20213d}
Richard~T Marriott, Sami Romdhani, and Liming Chen.
\newblock A 3d gan for improved large-pose facial recognition.
\newblock In {\em Proceedings of the IEEE/CVF Conference on Computer Vision and
  Pattern Recognition}, pages 13445--13455, 2021.

\bibitem{mescheder2018training}
Lars Mescheder, Andreas Geiger, and Sebastian Nowozin.
\newblock Which training methods for gans do actually converge?, 2018.

\bibitem{naik2021shallow}
Ankita Naik, Apurva Swarnakar, and Kartik Mittal.
\newblock Shallow-uwnet: Compressed model for underwater image enhancement
  (student abstract).
\newblock In {\em Proceedings of the AAAI Conference on Artificial
  Intelligence}, volume~35, pages 15853--15854, 2021.

\bibitem{Silberman:ECCV12}
Pushmeet~Kohli Nathan~Silberman, Derek~Hoiem and Rob Fergus.
\newblock Indoor segmentation and support inference from rgbd images.
\newblock In {\em ECCV}, 2012.

\bibitem{press2020emerging}
Ori Press, Tomer Galanti, Sagie Benaim, and Lior Wolf.
\newblock Emerging disentanglement in auto-encoder based unsupervised image
  content transfer.
\newblock {\em arXiv preprint arXiv:2001.05017}, 2020.

\bibitem{rahman1996multi}
Zia-ur Rahman, Daniel~J Jobson, and Glenn~A Woodell.
\newblock Multi-scale retinex for color image enhancement.
\newblock In {\em Proceedings of 3rd IEEE international conference on image
  processing}, volume~3, pages 1003--1006. IEEE, 1996.

\bibitem{royer2020xgan}
Am{\'e}lie Royer, Konstantinos Bousmalis, Stephan Gouws, Fred Bertsch, Inbar
  Mosseri, Forrester Cole, and Kevin Murphy.
\newblock Xgan: Unsupervised image-to-image translation for many-to-many
  mappings.
\newblock In {\em Domain Adaptation for Visual Understanding}, pages 33--49.
  Springer, 2020.

\bibitem{sarkar2021humangan}
Kripasindhu Sarkar, Lingjie Liu, Vladislav Golyanik, and Christian Theobalt.
\newblock Humangan: A generative model of human images.
\newblock In {\em 2021 International Conference on 3D Vision (3DV)}, pages
  258--267. IEEE, 2021.

\bibitem{simonyan2014very}
Karen Simonyan and Andrew Zisserman.
\newblock Very deep convolutional networks for large-scale image recognition.
\newblock {\em arXiv preprint arXiv:1409.1556}, 2014.

\bibitem{srivastava2017veegan}
Akash Srivastava, Lazar Valkov, Chris Russell, Michael~U Gutmann, and Charles
  Sutton.
\newblock Veegan: Reducing mode collapse in gans using implicit variational
  learning.
\newblock {\em Advances in neural information processing systems}, 30, 2017.

\bibitem{wang2021uwgan}
Nan Wang, Yabin Zhou, Fenglei Han, Haitao Zhu, and Jingzheng Yao.
\newblock Uwgan: Underwater gan for real-world underwater color restoration and
  dehazing, 2021.

\bibitem{wu2021novel}
Zhiheng Wu, Zhengxing Wu, Yue Lu, Jian Wang, and Junzhi Yu.
\newblock A novel underwater image synthesis method based on a pixel-level
  self-supervised training strategy.
\newblock In {\em 2021 IEEE International Conference on Real-time Computing and
  Robotics (RCAR)}, pages 1254--1259. IEEE, 2021.

\bibitem{Zhu_2017_ICCV}
Jun-Yan Zhu, Taesung Park, Phillip Isola, and Alexei~A. Efros.
\newblock Unpaired image-to-image translation using cycle-consistent
  adversarial networks.
\newblock In {\em Proceedings of the IEEE International Conference on Computer
  Vision (ICCV)}, Oct 2017.

\end{thebibliography}
}

\newpage

\begin{appendices}
\begin{figure*}[h]
\centering 
\includegraphics[width=17.3cm]{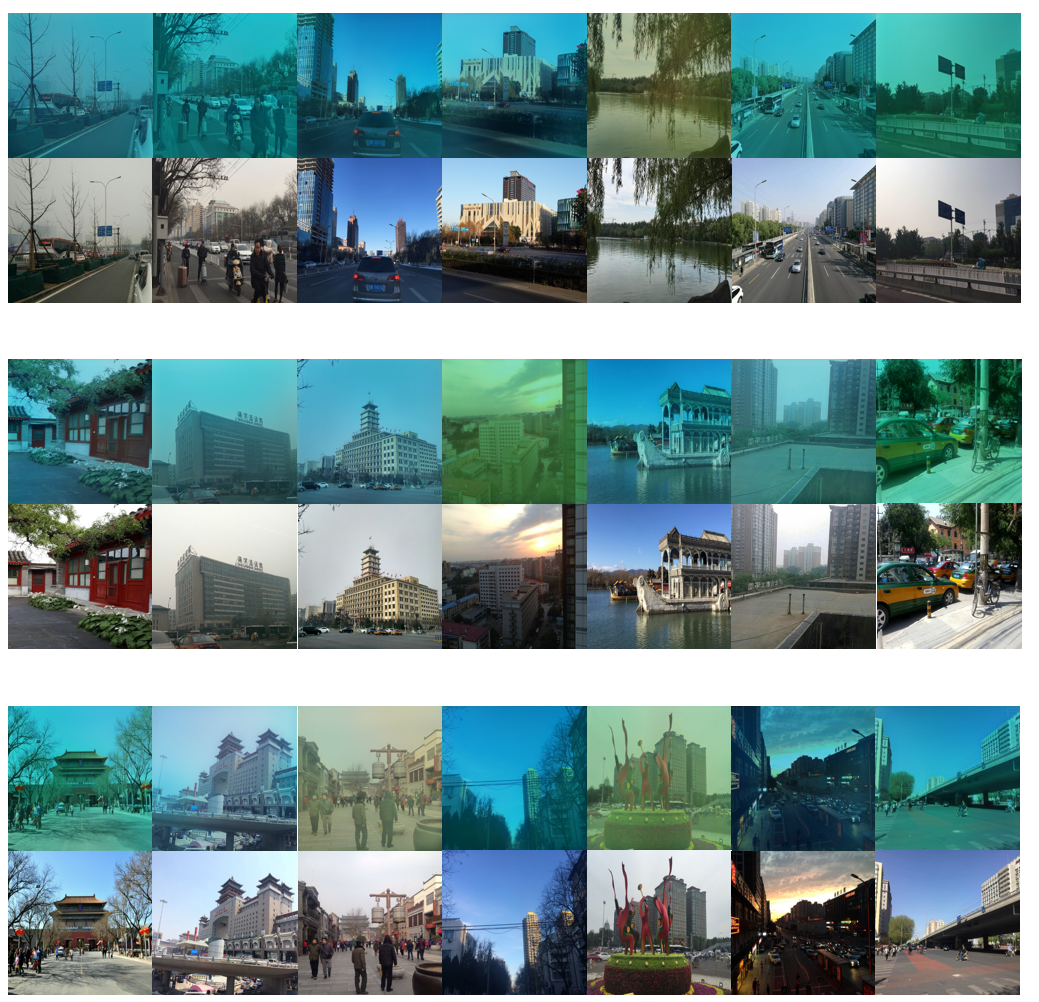}
\caption{Sampling images from our LNRU dataset. For each paired group, top: underwater image generated by our method, bottom: clean image. These clean images are from RESIDE~\cite{li2019benchmarking}. }
\label{fig:2}
\end{figure*}

\begin{figure*}[h]

\centering 
\includegraphics[width=17.3cm]{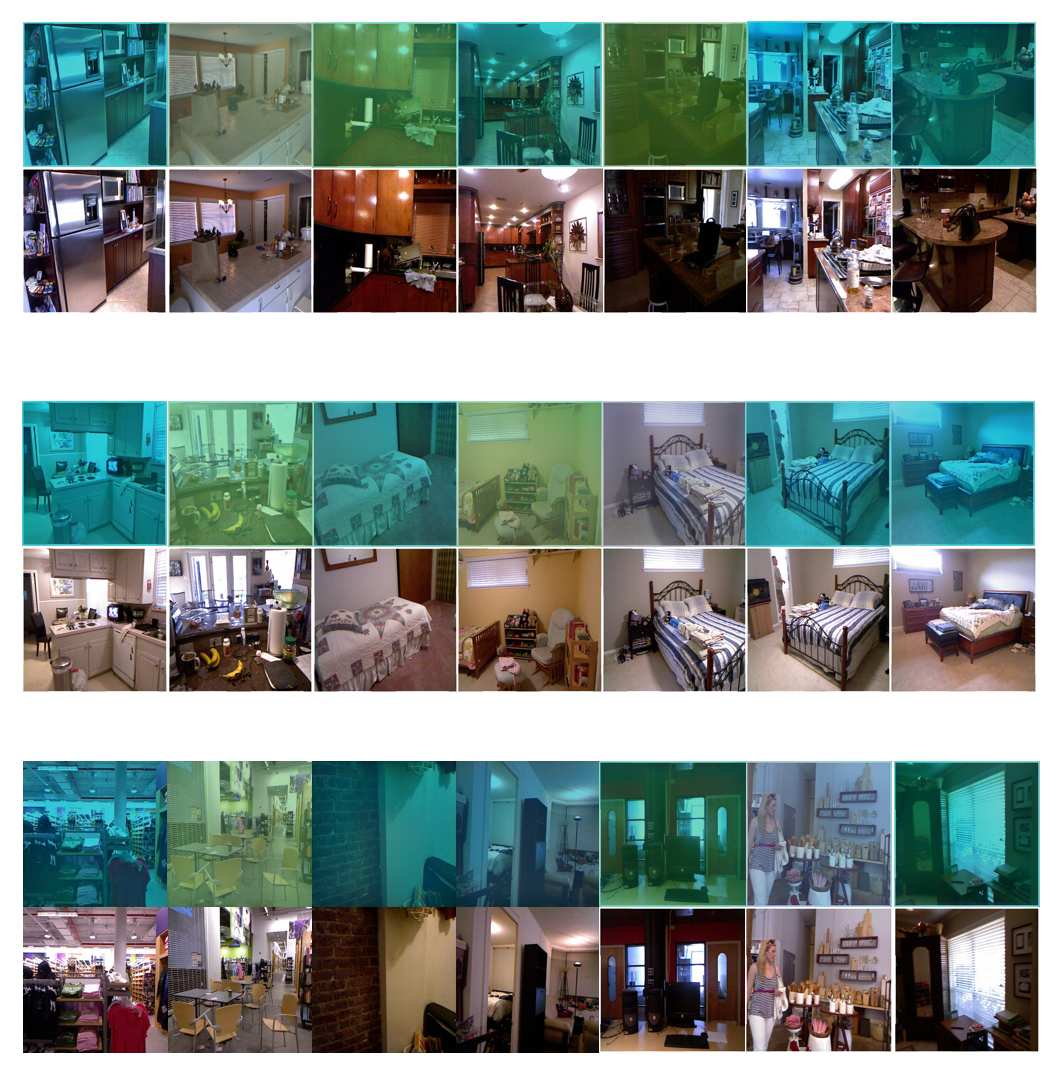}
\caption{Sampling images from our LNRU dataset. For each paired group, top: underwater image generated by our method, bottom: clean image. These clean images are from RESIDE~\cite{li2019benchmarking} and NYU~\cite{Silberman:ECCV12}.}
\label{fig:3}
\end{figure*}
\end{appendices}
%%%%%%%%% REFERENCES

\end{document}